\documentclass[runningheads]{llncs}
\usepackage[utf8]{inputenc}
\usepackage[T1]{fontenc}
\usepackage{graphicx}
\usepackage{todonotes}
\usepackage{amsmath}
\usepackage{amssymb}
\usepackage[toc,page]{appendix}

\usepackage{blindtext}

\usepackage{booktabs}
\newcommand{\otoprule}{\midrule[\heavyrulewidth]}
\usepackage{pifont}
\newcommand{\cmark}{\ding{51}}%
\newcommand{\xmark}{\ding{55}}%
\newcommand{\ubold}[1]{\fontseries{b}\selectfont#1}


\begin{document}
%

\title{ CASHformer: Cognition Aware SHape Transformer for Longitudinal Analysis}
%
%
\author{Ignacio Sarasua$^{1,2}$\thanks{Corresponding author: ignacio@ai-med.de}, Sebastian P{\"{o}}lsterl$^{2}$, Christian Wachinger$^{1,2}$}
\authorrunning{I. Sarasua, S. P{\"{o}}lsterl, C. Wachinger}
%
\institute{%
    $^{1}$Technical University of Munich, School of Medicine \\
$^{2}$Lab for Artificial Intelligence in Medical Imaging (AI-Med), KJP, LMU Klinikum 
}

\maketitle              

\begin{abstract}
Modeling temporal changes in subcortical structures is crucial for a better understanding of the progression of Alzheimer's disease (AD). Given their flexibility to adapt to heterogeneous sequence lengths, mesh-based transformer architectures have been proposed in the past for predicting hippocampus deformations across time. However, one of the main limitations of transformers is the large amount of trainable parameters, which makes the application on small datasets very challenging. In addition, current methods do not include relevant non-image information that can help  to identify AD-related patterns in the progression.
To this end, we introduce CASHformer, a transformer-based framework to model longitudinal shape trajectories in AD. 
CASHformer incorporates the idea of pre-trained transformers as universal compute engines that generalize across a wide range of tasks by freezing most layers during fine-tuning. 
This reduces the number of parameters by over 90\% with respect to the original model and therefore enables the application of large models on small datasets without overfitting.
In addition, CASHformer models cognitive decline to reveal AD atrophy patterns in the temporal sequence. 
Our results show that CASHformer reduces the reconstruction error by $73\%$ compared to previously proposed methods. 
Moreover, the accuracy of detecting patients progressing to AD increases by $3\%$ with imputing missing longitudinal shape data. 





\end{abstract}

\section{Introduction}

Alzheimer's disease (AD) is a complex neurodegenerative disorder that is characterized by progressive atrophy in the brain~\cite{Jack2013}. 
A spatio-temporal model of neuroanatomical changes is instrumental for understanding atrophy patterns and predicting patient-specific trajectories. 
For inferring such a model, longitudinal neuroimaging data can be used, but they are usually highly irregular with non-uniform follow-up visits and dropouts. 
Transformers provide a  flexible approach that can incorporate different sequence length inputs and incomplete time series. Hence, they are well suited for modeling longitudinal neuroimaging data.  
In addition, recent work has combined geometric deep learning on anatomical meshes with transformers to obtain a model that is sensitive to small shape changes in the hippocampus \cite{sarasua2021transformesh}. 

However, one of the main limitations of transformers is their large number of parameters.
Their huge success in Natural Language Processing (NLP) \cite{Devlin2018-bert} and Computer Vision \cite{dosovitskiy2020-vit} is also based on the availability of large datasets in these domains.
Medical datasets are typically much smaller, especially from longitudinal studies. 
This limits the application of transformers, particularly because deeper transformer networks are thought to be preferred \cite{li2020train}. 
At the same time, recent research established pre-trained transformers as \emph{universal compute engines} \cite{lu2021pretrained} that generalize across domains, e.g., pre-training on NLP and fine-tuning on tasks like numerical computation and vision. 
Based on this seminal work, we investigate whether transformers -- pre-trained on non-medical applications -- are helpful for creating a spatio-temporal model of progression to AD. 

Related research in generative shape modeling suggests that adding prior information (e.g., diagnosis) reduces the reconstruction error \cite{gutierrez2019learning}. 
However, in longitudinal modeling, diagnosis might remain the same along a patient's trajectory. 
Fortunately, the AD Assessment Scale (ADAS; \cite{mohs1997development}) cognitive score presents a fine-grained measure of cognitive decline along the longitudinal sequence. 
Including such information in a model can help aligning inter-patient data based on disease progression, as observed in \cite{mofrad2021cognitive}.

Given these considerations, we introduce CASHformer, a Cognition Aware SHape Transformer for longitudinal shape analysis. 
CASHformer generates embeddings from the input meshes of the hippocampus using a SpiralResNet \cite{azcona2021analyzing}. 
The embeddings are input into a transformer network, which has been pre-trained on a large non-medical dataset.
Following the Frozen Pre-Trained Transformer \cite{lu2021pretrained}, only a few layers are fine-tuned, which allows us to train deeper transformer architectures, while keeping the number of trainable parameters small. 
For explicitly modeling cognitive decline, we introduce ADAS embeddings  and an ADAS cost function that acts as regularizer. 
Our experiments demonstrate that CASHformer reduces the reconstruction error by $73\%$ with respect to previously proposed methods and increases the AD-progressor classification accuracy by 3\% with imputing missing shapes. 

\begin{figure}[t]
	\centering
	\includegraphics[width=\linewidth]{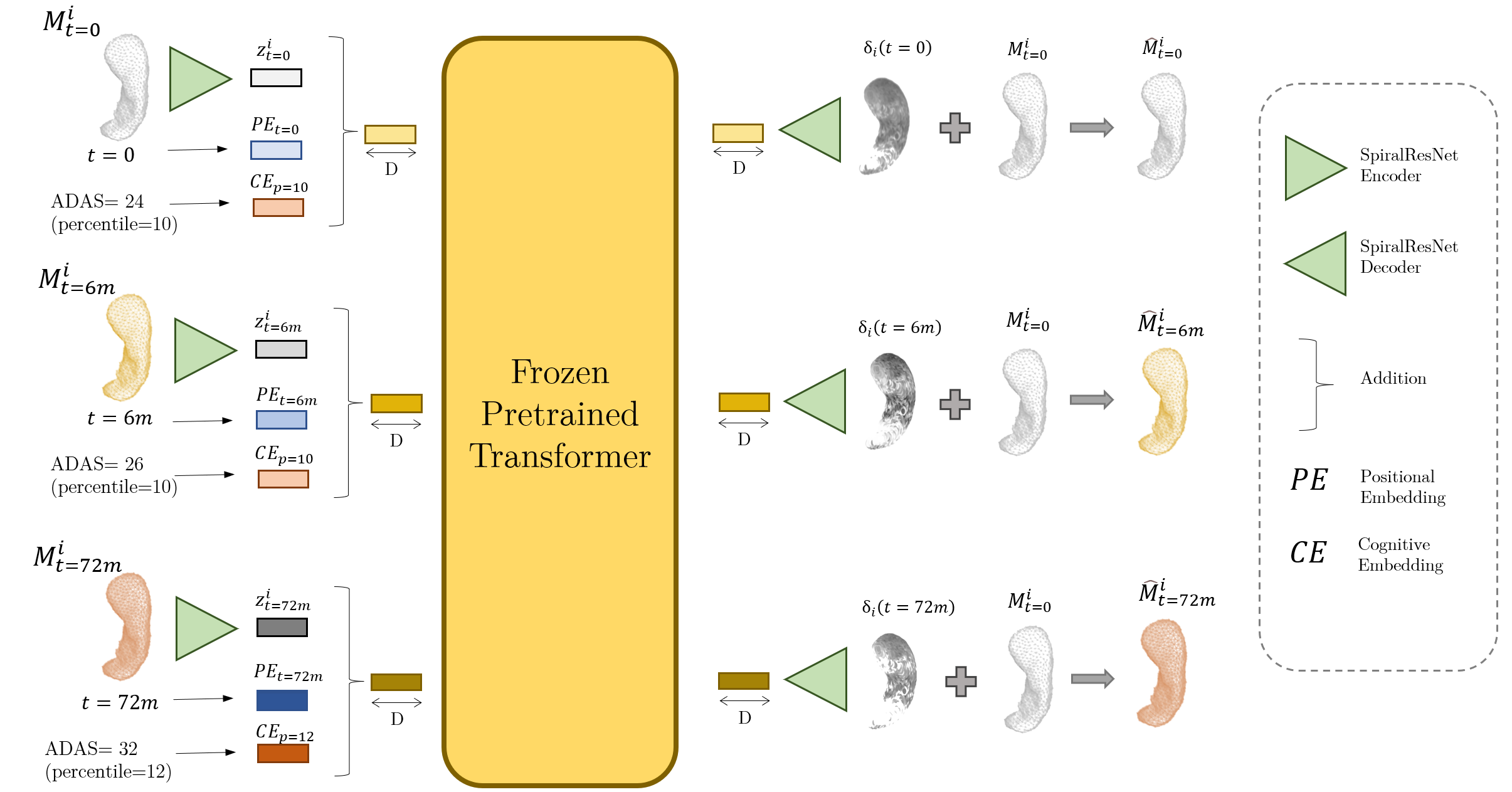}
	\caption{ Overview of the proposed CASHformer framework.
	}
	\label{fig:overview}
\end{figure}

\subsubsection{Related work.}
Long short-term memory (LSTM) networks have been previously applied to detecting AD, either given a single scan \cite{dua2020cnn,feng2019deep}, or a temporal sequence \cite{hong2019predicting}. Convolutional neural networks (CNNs) have been applied to longitudinal modeling of brain images in \cite{couronne2021longitudinal,zhao2021longitudinal}. These methods work on full brain images and cannot capture subtle changes in subcortical structures. In addition, they explicitly enforce linearity in latent representations within one patient. Our models take advantage of Multi-Head attention layers in transformers and positional embeddings to implicitly enforce this behavior. Given their capabilities of capturing subtle changes in brain structures, deep neural networks have been proposed to work on anatomical meshes \cite{azcona2021analyzing,sarasua2021geometric}.
Transformer networks \cite{Vaswani2017} have disrupted the fields of Natural Language Processing (NLP) \cite{Devlin2018-bert,Lewis2020-bart}, computer vision \cite{dosovitskiy2020-vit}, and medical image analysis~\cite{valanarasu2021medical,li2021few,yu2021mil}. Recently, they have also been applied to study anatomical structures like the hippocampus \cite{sarasua2021transformesh}. While the latter is closely related to ours, we propose to include cognitive performance as prior information, and a more efficient training strategy that allows us to train deeper transformer architectures with limited data.




\section{Methods}\label{sec:method}

Figure \ref{fig:overview} illustrates our proposed CASHformer framework. 
Let $\{M_t^{i} \,|\, t=0,...,T_i\}$ be the set of hippocampi meshes of subject $i$ from the time of enrollment $t=0$ to the last visit $t=T_i$.
The \emph{SpiralResNet encoder} generates an embedding for every available mesh $\{z_t^{i} \,|\, t=0,...,T_i\}$ of dimension $D$. The latent representations are then modulated by a set of learnable positional encodings and \emph{cognitive score embeddings}.
The former keeps track of the ordering of the sequence \cite{dosovitskiy2020-vit}, while the latter adds information about the cognitive evolution of the patient.
These latent representations are fed into a \emph{pre-trained transformer} and decoded by a \emph{SpiralResNet decoder}, yieling the deformation fields $\{\delta_i(t)\,|\, t=0,...,T_i\}$ that are applied to a reference shape. The model is trained end-to-end using a combination of reconstruction loss and \emph{cognitive aware loss}.

\subsection{Mesh Network}
\label{sec:mesh_network}
Since meshes do not follow a grid-like structure, operations like convolution or pooling are not as straightforward as for images. Therefore, we need a method that can define convolutional and down/up-sampling operations for such structures. Let $V_i$ be a vertex of a triangular mesh and $S(i,l)$ an
ordered set consisting of $l$ neighboring vertices of $V_i$, that are inside a spiral defined by choosing an arbitrary direction in counter-clockwise manner. 
The \emph{spiral convolution operation} for layer $k$ for features $\mathbf{x}_i$ associated to the $i$-th vertex is defined as $
    \mathbf{x}_i^{(k)} = \gamma^{(k)}(\parallel_{j \in S(i,l)} (\mathbf{x}_j^{(k-1)})),
$ where $\gamma$ denotes Multi-Layer Perceptrons (MLP) and $\parallel$ is the concatenation operation\cite{gong2019spiralnet++}. 
We work with a residual version  \cite{azcona2021analyzing}, where multiple SpiralConvolution layers  are concatenated with ELU and batchnorm layers, and residual connections are added (see Fig.~\ref{fig:mesh_network} in the supplemental material). 
For down-sampling the mesh after each convolution block, vertex pairs with the lowest quadratic error are contracted \cite{garland1997surface}.  
For up-sampling the meshes, the barycentric coordinates of the closest triangle in the downsampled mesh are used \cite{ranjan2018generating} (supplemental Fig.~\ref{fig:mesh_pool}). Given that all our shapes are registered to a template (see more in section \ref{sec:data_processing}), we can precompute the coordinates of the down-sampled and up-sampled meshes once for the template shape and then apply it to every patient in the dataset.

\subsection{Frozen Pre-trained Transformer}\label{sec:transformer}
Transformers are formed by a set of encoder blocks, where each of them consists of a Multi-Head-Attention and an MLP, preceded by a Layer Normalization (LN) block \cite{Vaswani2017}. 
In addition, residual connections are added after every block and GELU is used as non-linearity in the MLP block. 
The input is a sequence of embeddings, which in case of NLP are generated through tokenization methods that convert words into feature vectors like Byte Pair Encoding(BPE) \cite{radford2019language}. 
Subsequently, transformers were applied to image recognition tasks, by passing patches of the input image through a convolutional neural network to generate the embeddings, which achieved state-of-the-art performance on ImageNet \cite{dosovitskiy2020-vit}. 
However, one of the main issues of these architectures is their large number of trainable parameters. While this feature allows transformers to perform great on large datasets, it limits their application for smaller ones. 
 
Inspired by the Frozen Pre-Trained Transformer \cite{lu2021pretrained}, we use a pre-trained model  and only fine-tune the LN blocks.
This not only reduces the amount of trainable parameters by 90\%, but also allows us to apply larger transformer models (e.g., 24 blocks) to our smaller dataset \cite{radford2019language}.
To generate the embeddings that are the input to the transformer, we substitute the input layers of the transformers (i.e., CNN for ViT, and  BPE tokenizer for GPT-2) by a SpiralResNet, described in section \ref{sec:mesh_network}.


 \subsection{Cognitive Embeddings and Cognitive Decline Aware loss}
 The Alzheimer’s Disease Assessment Scale (ADAS) cognitive score is a widely used measure to study the progression of cognitive impairment \cite{mohs1997development}.
 Therefore, we introduce a set of learnable Cognitive Embeddings (CE) that modulate 
 the input latent representations coming from the mesh network (similar as the positional encoding ones \cite{dosovitskiy2020-vit}).
 The ADAS score ranges from 0 to 85.
 Since generating 85 embeddings would unnecessarily increase the number of trainable parameters, we quantize the whole range in 20 intervals based on the percentile they occupy within the training set (e.g., those shapes corresponding to ADAS score between percentile 10 and 15 are modulated using the same embedding). This results in a set of 20 embeddings of size $D$.
 The CE and mesh embedding are then summed to form a combined embedding (see Fig.~\ref{fig:overview}).

 Patients that experience acute cognitive decline also show larger deformation in their hippocampus \cite{mofrad2021cognitive}.
 We incorporate this clinical knowledge by introducing a regularization term, the Cognitive Decline Aware (CDA) loss.
 The CDA loss enforces those patients experiencing a larger cognitive decline to have a larger norm of their deformation field. 
 To this end, we maximize the cosine similarity (CS) between the line describing the ADAS score evolution and the sequence of the deformation field's norms $\mathcal{L}_\text{CDA}(\delta(t),ADAS(t))=\operatorname{CS}(\mathbf{\Delta}_{\delta(t)},\mathbf{\Delta}_{ADAS(t)})$,
 where $\mathbf{\Delta}_{\delta(t)} = \big[\|\delta(t=0)\|,\dots,\|\delta(t=T)\|\big]^\top$, $\mathbf{\Delta}_{ADAS(t)} = m \cdot [0,...,T]^\top$, and $m=(ADAS(T) - ADAS(0))/T$.

\subsection{Training procedure \label{sec:training}}
As illustrated in Figure \ref{fig:overview}, CASHformer predicts the mesh deformation field $\delta_i(t)$ with respect to a \emph{reference mesh}, which is a person's baseline shape $M_{t=0}^i$.

\noindent
\textbf{Missing shapes: } Transformer networks use padding tokens to fill up the positions of missing inputs.
These are ignored during training by adding key masks \cite{Vaswani2017}. 
However, the model still predicts an output for those positions,which  allows us to impute missing values during inference. For our application, we use the encoding of the reference shape, as produced by the mesh network, to generate the ``missing value'' embedding. As for the CE, we train an extra embedding to account for missing shapes.

\noindent
\textbf{Data Augmentation: } Inspired by \cite{Devlin2018-bert}, we apply random masking and shuffling to the input embeddings. In particular, 35\% of the shapes' embeddings are substituted by the reference shape's embedding and 15\% are shuffled along the sequence. Contrary to the real missing shapes, these are not ignored during training; the self-attention masks do take them into consideration.

\noindent
\textbf{Loss function: }The model is trained using a combination of Mean Square Error (MSE) and the CDA loss:
\begin{equation}\textstyle
    \mathcal{L} = \sum_{i=0}^{P} \sum_{t=0}^{T_i}||M^i_{t=0} + \delta_i(t) - M^i_{t}||^2 - \lambda \mathcal{L}_\text{CDA}(\delta_i(t),ADAS_i(t)), 
\end{equation}    
where $\lambda$ (empirically set to $10^{-6}$)  controls the contribution of the CDA loss, and $P$ is the number of patients in the set.

\section{Experiments}
\subsection{Data processing}
\label{sec:data_processing}
Given its relevance in AD pathology~\cite{Jack2013}, we use the shape of the left hippocampus for all our experiments. Our data is a subset of the Alzheimer’s Disease Neuroimaging Initiative (ADNI) data (adni.loni.usc.edu)~\cite{Jack2008}.
Structural MRI scans are segmented with FIRST~\cite{Patenaude2011} from  the FSL Software Library, which provides  meshes for the segmented samples. 
FIRST segments the subcortical structures by registering the MRI scan to a reference template, creating voxel-wise correspondences (and therefore, also vertex-wise) between the template and every sample in the dataset. This template is used for all the pre-computations of the mesh network in Sec.~\ref{sec:mesh_network}. We limit the number of follow-ups to $8$ (from baseline to $T_{max}=72$ months).
Since our focus is on modeling cognitive decline, we only include patients that have been diagnosed with MCI or AD.
Our subset of the ADNI data consists of  845 patients split 70/10/20 (train/validation/test) following a data stratification strategy that accounts for age, sex and diagnosis, so they are represented in the same proportion in each set. The average number of follow-up scans per patient is $3.32$ and only 3.6\% of the patients have attended all the follow-up sessions. 

\subsection{Implementation details}
For the mesh encoder and decoder, we followed \cite{azcona2021analyzing} autoencoder's architecture design (more details in supplemental Fig.~\ref{fig:mesh_network}).
For the transformer, we evaluated  three state-the-art pre-trained transformer architectures: GPT-2 \cite{radford2019language} trained on the WebText dataset, ViT$_{base}$(12 blocks) and ViT$_{large}$(24 blocks) trained on ImageNet-21k and fine-tuned on ImageNet-1k\footnote{https://github.com/rwightman/pytorch-image-models}.
As baselines, we compare to TransforMesh \cite{sarasua2021transformesh}, the substitution of the transformer part (including the positional and cognitive embeddings) by LSTM networks\footnote{https://pytorch.org/docs/stable/generated/torch.nn.LSTM.html}, and training GPT-2 and ViT networks from scratch. 
Note that only TransforMesh has been used in mesh-based progression modeling before.
More details about the architecture design can be found in supplemental Table~\ref{tab:arch_details}. 

\subsection{Longitudinal shape modeling}
We evaluate each model on three types of tasks: shape interpolation,  shape extrapolation, and trajectory prediction. For each of these experiments, a set of input shapes, $M_{t}^i$, are removed for every patient. For a fairer comparison to the other methods, and to simulate a more realistic scenario where a patient missed the visit at time $t$, we do not input the ADAS score for that follow-up (i.e. we use the extra CE described in section \ref{sec:training}).
The corrupted sequence is passed through the network and the Mean Absolute Error (MAE) is computed between  $M_{t}^i$ and $\hat{M}_{t}^i$, where  $\hat{M}_{t}^i$ is the predicted mesh at visit $t$.

\noindent
\textbf{Interpolation:} From each patient in the test set, the shape in the middle of the input sequence, $M_{t=\mu}^i$ with $\mu = \lfloor T_i/2 \rfloor$, is removed. The goal of this experiment is to evaluate the capabilities of the models to predict a missing shape using both past and previous information. 

\noindent
\textbf{Extrapolation:} From every patient in the test set, we remove from the sequence the shape $M_{t=T_i}^i$ and input the remaining shapes to the network. This experiment aims to measure the performance of each model to predict the mesh of the last available visit, based on all the previous visits. 

\noindent
\textbf{Trajectory prediction:} The third experiment is similar to the extrapolation experiment, but it predicts shapes that are more distant in time. 
Therefore, we only input the shape belonging to the baseline scan $M^i_{t=0}$, and predict all the shapes that are at least 2 years apart ($24\text{m} \leq t \leq T_i$).


\begin{table}[tb]
\setlength{\tabcolsep}{2.5pt}
\scriptsize
\centering
\caption{\label{tab:results}%
Top: Median error and median absolute deviation for interpolation, extrapolation, and trajectory experiments are reported. Errors were multiplied by $10^3$ to facilitate presentation. Bottom: Evaluation of each of the contributions in CASHformer.}
\begin{tabular}{lccccr}
\otoprule
Method          &Pre-train        & Interpolation          & Extrapolation           & Trajectory    &Parameters   \\
\midrule
Small TransforMesh \cite{sarasua2021transformesh}  &\xmark     & $5.454 \pm 0.289$ & $5.573 \pm 0.317$ & $5.592 \pm 0.294$ & $10.4M$ \\
Base TransforMesh \cite{sarasua2021transformesh}  &\xmark   & $6.352 \pm 0.334$ & $6.589 \pm 0.381$ & $6.560 \pm 0.379$ & $38.7M$\\
LSTM(3 blocks)           &\xmark         & $6.610 \pm 0.060$ & $6.610 \pm 0.060$ & $6.832 \pm 0.341$ & $20.1M$\\
LSTM(12 blocks)    &\xmark & $5.263 \pm 0.094$ & $5.258 \pm 0.088$ & $5.653 \pm 0.308$ & $62.6M$\\
GPT-2 \cite{radford2019language}   &\xmark                  & $5.258 \pm 0.124$ & $5.267 \pm 0.127$ & $5.692 \pm 0.351$  & $130M$\\
ViT$_{base}$ \cite{dosovitskiy2020-vit}     &\xmark              & $3.713 \pm 0.076$ & $3.713 \pm 0.073$ & $4.040 \pm 0.262$ & $86.4M$\\
ViT$_{large}$ \cite{dosovitskiy2020-vit} &\xmark  & $ 6.514 \pm 0.290$ & $6.610 \pm 0.305$ & $6.709 \pm 0.286$ & $309M$\\
\midrule
CASHformer w/ GPT-2 &\cmark  & 6.111 $\pm$ 0.222 & 6.362 $\pm$ 0.324 & 6.353 $\pm$ 0.311 & $5.99M$\\
CASHformer w/ GPT-2 &\xmark  & 3.683 $\pm$ 0.223 & 3.953 $\pm$ 0.335 & 3.955 $\pm$ 0.324 & $130M$\\
CASHformer w/ ViT$_{base}$ &\cmark  & 3.200 $\pm$ 0.256 & 3.402 $\pm$ 0.348 & 3.418 $\pm$ 0.366 & $5.99M$\\
\textbf{CASHformer w/ ViT$_{large}$} &\cmark  & \ubold{1.472 $\pm$ 0.257} & \ubold{1.718 $\pm$ 0.337} & \ubold{1.801 $\pm$ 0.391} & $6.84M$\\
\bottomrule\\[-.6em]
\otoprule
Base Model in CASHformer                  & Pre-train           & CE     & CDA       & Interpolation & Parameters \\
\midrule
ViT$_{large}$  & \xmark & \xmark & \xmark & $ 6.514 \pm 0.290$ & $309M$\\
ViT$_{large}$  & \cmark & \xmark & \xmark & $2.610 \pm 0.149$ & $6.82M$\\
ViT$_{large}$  & \cmark & \cmark & \xmark & $1.623 \pm 0.261$ & $6.84M$\\
ViT$_{large}$ & \cmark & \cmark & \cmark & $1.472 \pm 0.257$ & $6.84M$\\
\bottomrule
\end{tabular}
\end{table}

\subsection{Results}
\subsubsection{Mesh prediction.}
Table \ref{tab:results} reports the median error and the median absolute deviation across all patients and visits. 
Regarding the baseline methods, ViT$_{base}$ performs best.
The second best performing baseline is the deeper version of LSTM (with 12 blocks). 
Compared to its more shallow counterpart with 3 blocks, LSTM12 reduces the error by approximately $20\%$, but triples the number of parameters.  
We observe the opposite behaviour for TransforMesh, where the Small TransforMesh performs better than the Base TransforMesh. Note that the TransforMesh architectures use the regular SpiralNet++ as their mesh backbone, while  the rest of the methods  use SpiralResNet. 
For CASHformer, ViT$_{large}$ leads to the best results with a reduction of the reconstruction error by 60\% w.r.t. the best performing baseline (ViT$_{base}$) and 73\% w.r.t. to previously proposed TransforMesh, while having 90\% and 33\% less trainable parameters, respectively.
Interestingly, the improvement of CASHformer w/ ViT$_{base}$ compared to ViT$_{large}$ is much smaller. 
These results are in line with prior studies about the superiority of deeper transformer architectures \cite{li2020train}.
The comparison between CASHformer with ViT and GPT-2 models shows that vision transformers are more suitable for our application. 
We believe that the distribution of the latent space in NLP models, which is generated from input embeddings following tokenization methods like BPE \cite{radford2019language}, is too different to our embeddings coming from a mesh network. 
For this reason, we also evaluated the effect of only using the CE and the CDA loss for GPT-2, without pre-training. We observe that adding these two features reduces the reconstruction error, w.r.t. using the original GPT-2, by $30\%$.

\begin{figure}[tb]
    \centering
    \includegraphics[width=.8\textwidth]{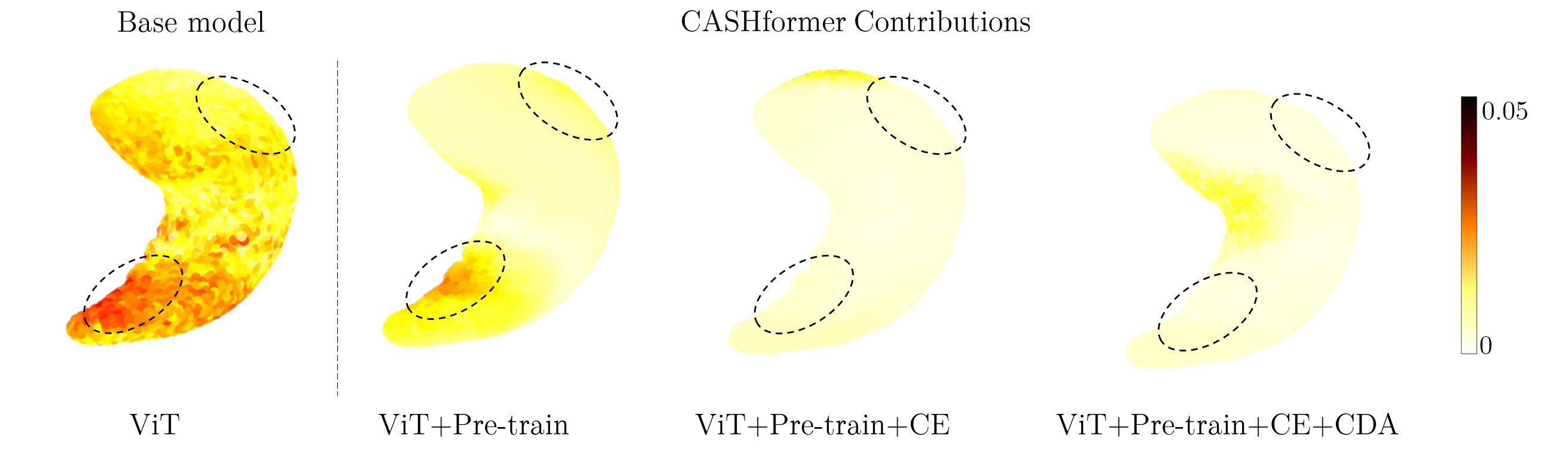}
    \caption{\label{fig:rec_error}%
    Hippocampus reconstruction error for the contributions of CASHformer. The highlighted areas correspond to  the medial part of the body in the subiculum  and parasubiculum areas, and  the lateral part of the body in the CA1 area.}
\end{figure}


Given the superior performance of ViT$_{large}$, we study the effect of each contribution in Table \ref{tab:results} and Fig.~\ref{fig:rec_error}. 
Fine-tuning the LN blocks  reduces the reconstruction error by $60\%$ with respect to the original model. 
Adding the CE yields a further reduction of the error by $38\%$, and adding the CDA loss by another $10\%$.
Fig.~\ref{fig:rec_error} shows the average reconstruction error along all  patients for the interpolation experiment.
We can observe that adding ADAS information reduces the reconstruction error in the medial part of the body in the subiculum  and parasubiculum areas, and the lateral part of the body of the CA1 area (highlighted in the figure).
These parts of the hippocampus have been found to suffer larger atrophy with progression of  dementia~\cite{lindberg2012shape}. Adding cognitive information drives more attention to those AD-related areas.  



\subsubsection{Trajectory classification.}
As final experiment, we evaluate the differentiation between MCI subjects that remain stable and those that progress to AD within the study period. 
Once we classify based on the original shape sequence, and once we classify based on imputing the missing shapes with CASHformer w/ ViT$_{large}$. 
As classifier, we fine-tune a separate ViT$_{base}$ model. 
The mean classification accuracy without imputation is 0.73 and with imputation is 0.76 ($t$-test $P=0.003$).
A boxplot of the results is in supplemental Fig.~\ref{fig:cls_boxplot}.
The increase of the classification accuracy by 3\% after imputation by CASHformer not only shows that our model can be used to boost performance, but it also confirms that it is able to learn meaningful representations that model the evolution of hippocampus atrophy in AD.


\section{Conclusion}
We have proposed CASHformer, a transformer-based framework for the longitudinal modeling of  neurodegenerative diseases. 
Our results demonstrated that pre-trained transformers on vision tasks can generalize to medical tasks, despite the large domain gap, supporting their role as universal compute engines. 
We believe that this opens up new avenues for using large transformers models on medical tasks, despite scarcity of data. 
Moreover, our results illustrated the importance of including cognition data in the model through CE and the CDA loss to focus on AD-related shape changes. 
Finally, we showed the capability of CASHformer to impute missing shapes for improving the discrimination between MCI subjects that remain stable and progress to AD.

\paragraph{Acknowledgment}
This research was partially supported by the Bavarian State Ministry of Science and the Arts and coordinated by the bidt, and the Federal Ministry of Education and Research in the call for Computational Life Sciences (DeepMentia, 031L0200A). We gratefully acknowledge the computational resources provided by the Leibniz Supercomputing Centre (www.lrz.de).


\bibliographystyle{splncs03.bst}
\bibliography{biblio.bib}

\clearpage
\appendix

\renewcommand\thefigure{S\arabic{figure}}%
\renewcommand\thetable{S\arabic{table}}%
\setcounter{figure}{0}%
\setcounter{table}{0}%

\begin{figure}[t]
    \centering
    \includegraphics[width=\textwidth]{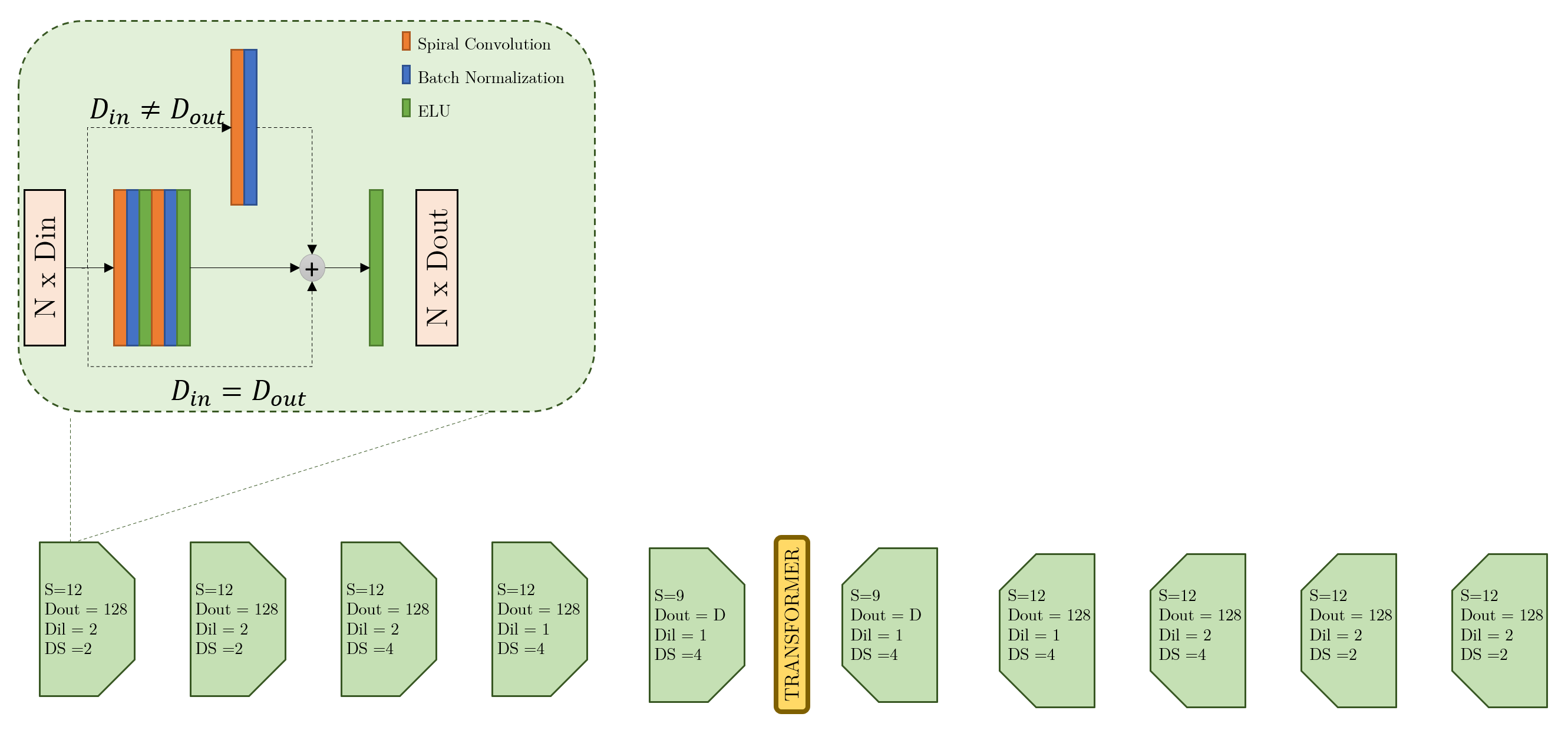}
    \caption{Top: Architecture of a ResSpiralBlock that constitutes the encoder and decoder. Bottom: Mesh encoder and decoder architecture details. (\textit{S}=length of sequence; \textit{Dout}: Number of output channels; \textit{Dil}: Dilation factor; \textit{DS}: Downsamling factor. \textit{Din}=Number of input channels) }
    \label{fig:mesh_network}
\end{figure}

\begin{figure}[t]
    \centering
    \includegraphics[width=\textwidth]{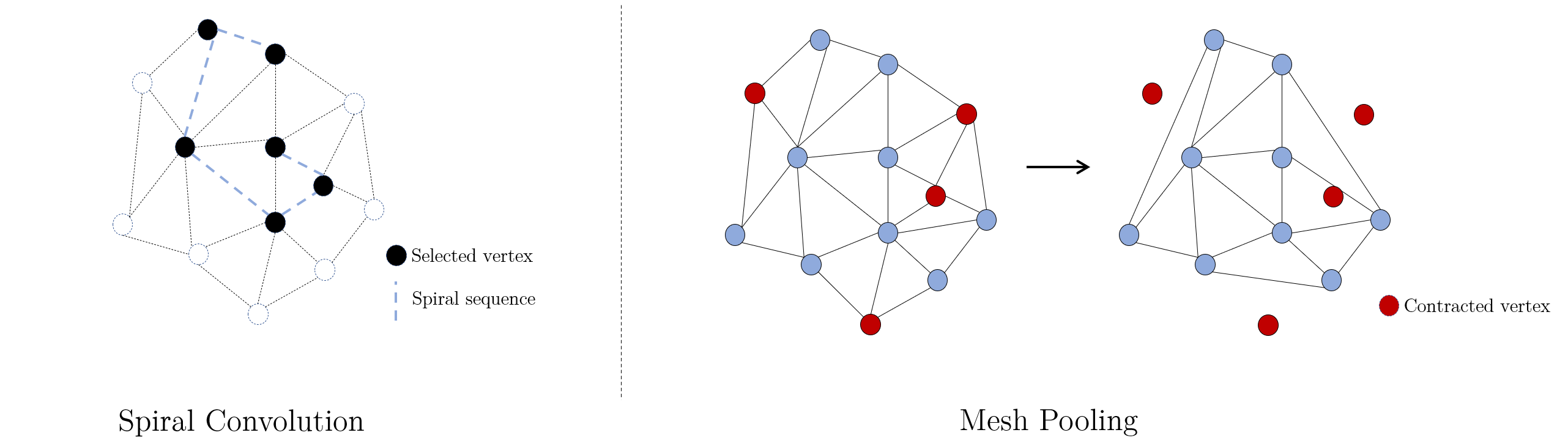}
    \caption{Left: example of spiral sequence for convolution. Right: Pooling operation by minimum quadric error (red vertices get contracted).}
    \label{fig:mesh_pool}
\end{figure}

\begin{figure}[t]
    \centering
    \includegraphics[width=\textwidth]{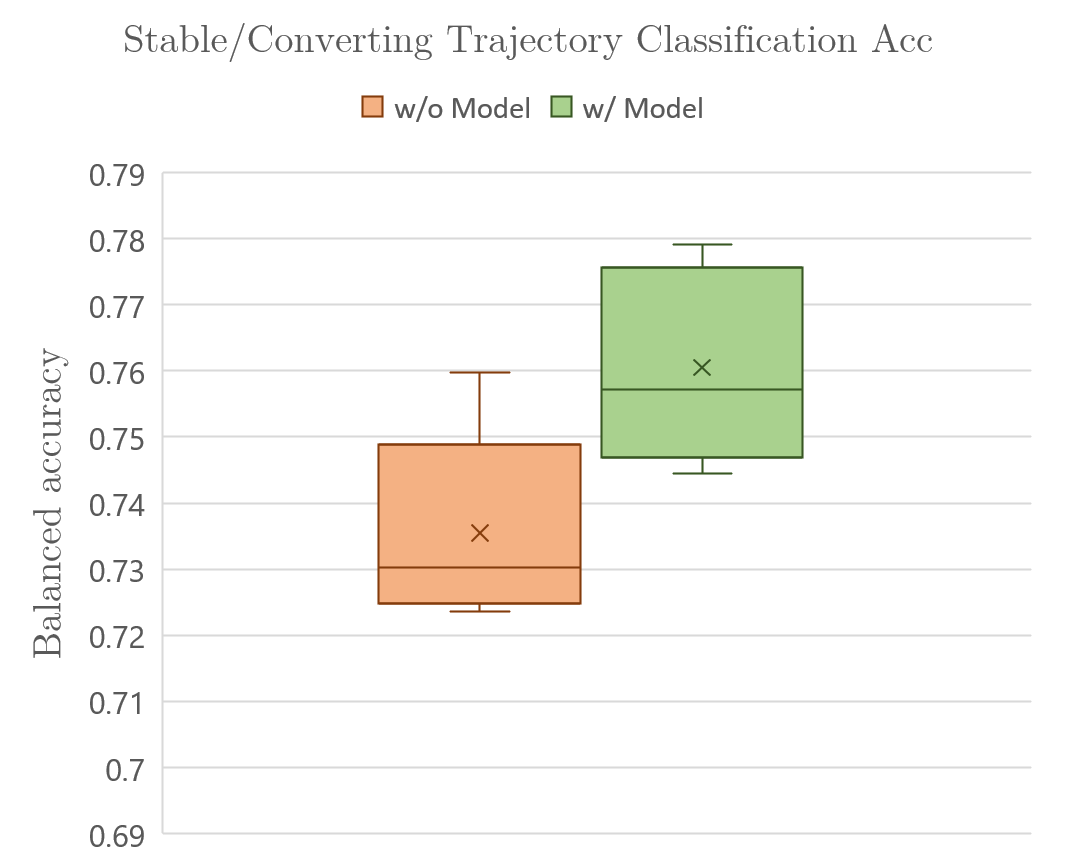}
    \caption{Balanced accuracy for classifying stable or converting trajectories. Difference between using the CASHformer+ViT$_{large}$ to impute the missing shapes (green) and using the original sequence (orange). }
    \label{fig:cls_boxplot}
\end{figure}

\begin{table}[t]
\centering
\caption{
\label{tab:arch_details}%
Transformer architecture details. All networks were trained during 1000 epochs on a single NVIDIA Titan Xp, with a constant learning rate of $10^{-3}$ and using Adam as the optimizer.  }
\begin{tabular}{lccc}
\otoprule
Model              & Number of blocks & Latent Dimension (D) & Hidden Size \\
\midrule
Small TransforMesh & 3                & 512                  & 2048        \\
Base TransforMesh  & 12               & 512                  & 2048        \\
LSTM$_3$           & 3                & 768                  & 768         \\
LSTM$_{12}$          & 12               & 768                  & 768         \\
GPT2               & 12               & 768                  & 3072        \\
ViT$_{base}$       & 12               & 768                  & 3072        \\
ViT$_{large}$      & 24               & 1024                 & 4096         \\
\bottomrule
\end{tabular}
\end{table}

\end{document}